# DATA-LEVEL HYBRID STRATEGY SELECTION FOR DISK FAULT PREDICTION MODEL BASED ON MULTIVARIATE GAN

Shuangshuang Yuan [1], Peng Wu[1] and Yuehui Chen [1(✉)]

[1] School of lnformation Science and Engineering, University of Jinan, Jinan, China
yuanshuang1024@163.com, ise_wup@ujn.edu.cn, yhchen@ujn.edu.cn

## ABSTRACT

*Data class imbalance is a common problem in classification problems, where minority class samples are often more important and more costly to misclassify in a classification task. Therefore, it is very important to solve the data class imbalance classification problem. The SMART dataset exhibits an evident class imbalance, comprising a substantial quantity of healthy samples and a comparatively limited number of defective samples. This dataset serves as a reliable indicator of the disc's health status. In this paper, we obtain the best balanced disk SMART dataset for a specific classification model by mixing and integrating the data synthesised by multivariate generative adversarial networks (GAN) to balance the disk SMART dataset at the data level; and combine it with genetic algorithms to obtain higher disk fault classification prediction accuracy on a specific classification model.*

## 1. INTRODUCTION

The basic storage device in the storage sector, the hard disk, is still the dominant device. To safeguard data integrity and prevent operational disruptions, it is vital to prioritize the implementation of disk failure prediction. This is primarily due to the inherent nature of disk damage, where retrieval of data becomes highly improbable once the disk is compromised.

There exists a substantial disparity between the number of both positive and negative data instances(positive for failed samples and negative for healthy samples) in the SMART dataset, which is employed for the purpose of disk failure prediction, and the number of positive samples is far less than the number of negative samples when analysed against the widely used Backblaze public dataset. The effects of error-prone small separations, overlapping classes and the lack of training data for classifiers make it more difficult to design effective ML models [1]. The majority of prevailing machine learning models operate under the premise that the distribution of both positive and negative data instances is relatively balanced, along with the assumption of a uniform misclassification cost. When using the more common and efficient machine learning models on datasets with a large proportion of imbalances, this can lead to training models that focus on the majority class and identify all sample data as the majority class [2], whereas the class with a small number of samples (faulty samples) is often more important and significant than the other classes of data. In the field of disk failure prediction, the most common method used by researchers to solve the classification problem caused by class imbalance is the data-level approach, i.e. pre-processing the imbalanced dataset to achieve class balance before training for classification.

Data-level processing methods for disk SMART data imbalance are classified as: oversampling, undersampling, hybrid sampling, and generating model-generated data. The simplest of the oversampling methods [3-5] is random oversampling, which reduces the rate of class imbalance by replicating a small number of class samples so that the number of samples from both classes is similar, thus allowing more balanced training of the classifier. However, increasing the number of minority class samples creates too much redundant and overlapping data, leading to overfitting of the classification model to the replicated samples. The Synthetic Minorit Oversampling Technique (SMOTE) [6], by analysing the minority class samples (real samples), randomly selecting samples of similar distance, and generating new minority class samples without duplicates by the difference between two samples, can overcome the overfitting problem of the random oversampling method to some extent, but the type and number of samples generated, as well as the ability of SMOTE to handle sample noise, need further improvement and optimisation. Undersampling methods [7-9] are used to effectively reduce the imbalance between the majority and minority classes by reducing the number of majority class data. Random undersampling selects a random portion of samples from the majority class so that the number of selected majority class samples is equal to the number of minority class samples to form a new class-balanced dataset, but the majority class samples are ignored in this process, which affects the diversity and representativeness of the data and poses a challenge for the classifier to correctly learn the decision boundary between both positive and negative data instances, thus reducing the classifier performance. Hybrid sampling methods balance a data set by combining different data sampling methods. A common hybrid sampling method is a combination of oversampling and undersampling methods, where a few classes of data are oversampled and most classes are undersampled.

By fitting the eigenfunctions of the samples using traditional methods, the simulated fault disc data can deviate significantly from actual faulty disc data, hence, these data are suboptimal. The above methods only change th0e number of data samples, but do not significantly enrich the sample set by increasing the number of samples in a few classes. By learning the probability distribution of authentic tabular data, Table GAN(a generic term for GAN models that can synthesise table data) can generate synthetic data that replicates the original distribution of authentic samples, thereby achieving the desired falsification outcome. The use of disk failure samples synthesised by Table GAN to balance the dataset [10,11] increases the diversity of the sample set while reducing the positive and negative sample imbalance ratio, which is beneficial to the improvement of disk failure prediction accuracy, but it is sometimes impossible to achieve Nash equilibrium using gradient descent. As a consequence, the training process of the GAN becomes precarious and occasionally fails to entirely grasp the distinctive attributes and underlying distribution of genuine data, resulting in the inability to generate synthetic data that accurately aligns with the distribution of authentic data.

The innovative points in this paper are outlined as follows:

In order to tackle the aforementioned issues, in this paper, we use multivariate GAN synthesized disk failure data before training the machine learning model, and find the optimal mixing ratio of the synthesized data by optimization algorithm (GA). The approach of this paper involves adding synthesized data from a multivariate GAN to the real dataset. This is done by determining the optimal mixing ratio, resulting in a disk SMART dataset that possesses a balance between both positive and negative data instances. The experiments are validated by repeated experiments on classification models such as Multi-Layer Perceptron (MLP), Support Vector Machine (SVM), Decision Tree (Dec_Tree), Bayesian Network (Bayes) and RandomForest. According to the experimental findings, the optimal data is discovered by integrating the genetic algorithm with a hybrid integration approach that combines multivariate GAN-synthesized data mixing ratio can construct a balanced data set suitable for a specific classification model, which can better alleviate the imbalance between positive and negative samples of disk SMART data and improve the accuracy of disk failure prediction.

## 2. RELATED WORK

## 2.1. Generative Model

### 2.1.1. CTGAN

CTGAN, known as "Conditional Tabular GAN", is an extension of GAN proposed by Xu L [12] at NeurIPS 2020. Its innovation lies in its ability to model the probability distribution of rows (samples) in tabular data and to generate realistic synthetic data. That is, CTGAN can model the distribution of features of continuous and discrete columns contained in the tabular data to generate synthetic data sets that match the distribution of the original data.

For continuous columns, CTGAN generates new data based on the selected probability distribution function. At the same time, CTGAN generates more robust data by adding noise to the continuous columns according to the method used in the algorithm to protect the identifier information. For the category columns, CTGAN adds a conditional vector to the generator that fully captures the distributional characteristics of the category columns. Therefore, CTGAN is able to simulate the category proportions in the dataset very well at the time of generation, thus generating category columns that match the conditional distribution of the original dataset; and solving the data category imbalance problem by employing conditional generators and implementing training through sampling techniques.

### 2.1.2. CopulaGAN

To generate tabular datasets, CopulaGAN [13] employs a CDF-based transformation facilitated by Gaussian copulas as an adaptation of CTGAN. It leverages probabilistic integral transformations [14] to improve the fitting of the CTGAN model. CopulaGAN uses a copula function-based approach to learn and model the distributional relationships between arbitrary columns to capture the interactions between continuous and discrete columns in a dataset. Each column is modelled with a different Copula function in CopulaGAN and trained by the GAN framework to make the Copula distribution more similar to the true distribution of the dataset. The generator can then use the generated copula distributions to generate new tabular datasets while maintaining the correlation between all columns in the data.

Unlike traditional GANs, CopulaGAN does not use random noise as input, but specifies input conditions, such as the distribution conditions of continuous and discrete columns. As a result, CopulaGAN can generate synthetic data sets that meet the requirements based on different input conditions, and can produce data sets of different sizes to meet different needs.

### 2.1.3. CTAB-GAN

The CTAB-GAN [15] model can effectively model a variety of data types, including continuous and discrete variables, and addresses data imbalance and long-tail problems in real tabular data sets by exploiting the information loss and classification loss of conditional GAN. The model has special conditional vectors that can effectively encode mixed data types and skewed distributions of data variables.

CTAB-GAN has the following advantages: large generation size: CTAB-GAN's tabular data generator can generate synthetic data sets of arbitrary size, which can be adapted to the needs of data of different sizes; high interpretability: CTAB-GAN's generator can not only take random noise as input, but can also specify input conditions. Thus, the generated data can be controlled by the input conditions, which is highly interpretable. Support for multi-attribute data generation: CTAB-GAN can model relationships between any number and type of columns to generate tabular data with multiple attributes. Data privacy protection: CTAB-GAN generates data without storing the original data, which protects data privacy.

## 2.2. Classification model

### 2.2.1. MLP

MLP, also known as Artificial Neural Network (ANN), is a forward-structured neural network that can be used to solve a variety of practical problems, such as classification problems, regression problems, image processing, and so on. It has good non-linear adaptation and generalisation capabilities. MLP consists of three layers: an input layer, a hidden layer and an output layer, where the nodes in each layer are fully connected to the nodes in the next layer. The input layer receives the input, the hidden layer processes the input, and the output layer generates the result. In addition to the input and output layers, it can have several hidden layers in between, as shown in Figure.1 below.

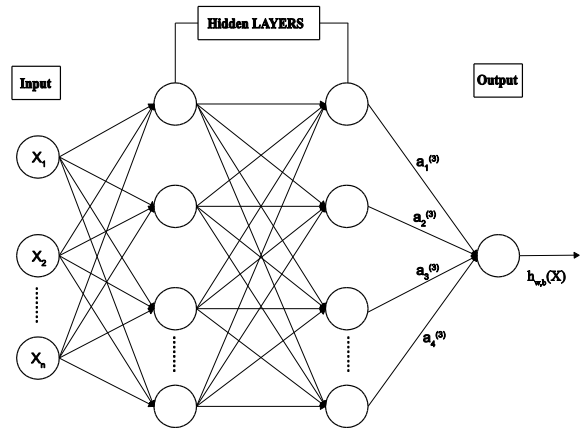

Figure 1. MLP network structure

MLP layers are fully interconnected, with the bottom layer being the input layer, the middle layer being the 2 hidden layers, and the top layer being the output layer. Due to its very good nonlinear mapping capability, high parallelism and global optimisation, MLP has achieved good results in image processing, prediction systems and pattern recognition today [16].

### 2.2.2. SVM

Support vector machine is a machine learning algorithm based on the principle of structural risk minimisation, which is based on statistical learning theory and automatically adjusts the model structure by controlling parameters to achieve empirical and structural risk minimisation [17]. For non-linear problems, such as disk failure prediction problems, SVM uses kernel functions to map the input data into a high-dimensional space to achieve linear separability of the high-dimensional space, thus transforming non-linear problems into linear problems.

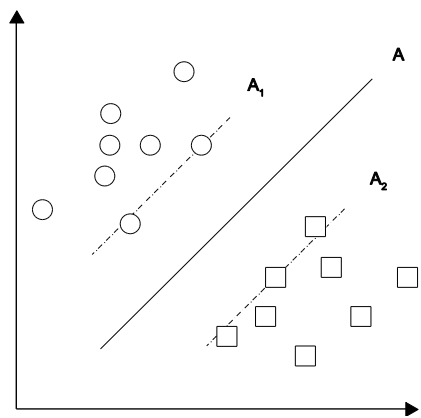

Figure 2. Support vector machine classification hyperplane

As shown in Figure.2, the left and right sides of the diagram represent different types of data, A, A1, A2 are the classification surfaces of the two types of data, where A1, A2 is the edge classification surface that divides the two types of data, margin is the classification interval between the two types, and the data points on A1, A2 is the support vector. The purpose of the support vector machine is to find an optimal classification surface that maximizes the classification interval between the two classes [18].

### 2.2.3. Decision Tree

A decision tree is a tree-based machine learning algorithm that is widely used in prediction models for split problems. It uses a tree structure to extract multiple decision rules and gradually branches down from the root node to a leaf node, resulting in a final prediction [19]. In a decision tree, each internal node represents a decision to examine feature values, every leaf node symbolizes either a specific category or a prediction. The data set is recursively partitioned and divided to generate a decision tree. The implementation of the decision tree is as show in algorithm 1.

```
Algorithm 1 Algorithm of decision Tree
Input: Training set D={ (x_1,y_1)(x_2,y_2),....(x_m,y_m) }; Attribute set A = { a_1, a_2, ....a_d },
Output: An existing decision tree with node as the root node
 1: Generating node: node
 2: if the samples in D are all in the same category C then
 3:     Mark node as a class C leaf node; return
 4: end if
 5: if {A = ∅} OR The samples in D are the same on A then
 6:     Mark node as a leaf node and its category as the class with the largest sample size in
        D; return
 7: end if
 8: Select the optimal partition attribute A from a_*;
 9: for Every value a_*^v in a_* do
10:     Generate a branch for node; Let D_v represent D represent the sample subset of a_*^v on
        a_*;
11:     if D_v is NULL then
12:         The branch node is marked as a leaf node, and its type is marked as the class with
            the most samples in D; return
13:     else
14:         Take TreeGenerate(D_v, A\{a_*}) as the branch node
15:     end if
16: end for
```

### 2.2.4. RandomForest

Random forest, a representative algorithm of the integrated learning bagging method, is a comprehensive learning approach grounded on decision trees. The core idea is to use the bagging method to build multiple decision tree models for multiple sets of returned samples, specifically: each decision tree is a classifier and for an input sample, N trees yield N classification results, while Random Forest integrates all classification votes and designates the category with the most votes as the final output. In this method, each decision tree is constructed from randomly selected samples from the training set, the random forest randomly selects a subset of features to randomise the tree, and the features selected during the construction of the decision tree and the features of the split nodes are randomised. This method has a high degree of randomness, which can prevent overfitting and improve the performance of the algorithm.

### 2.2.5. GaussianNB

It is assumed that the conditional probability (Equation follows) of each characteristic variable under each category follows the Gaussian distribution, and then the posterior probability of a new sample belonging to each category under a given characteristic distribution is calculated according to the Bayesian formula.

$$P(x_i|y_c) = \frac{1}{\sqrt{2\pi\sigma_{ci}^2}} \exp\left(-\frac{(x_i - \mu_{ci})^2}{2\sigma_{ci}^2}\right)$$

Where $x_i$ denotes the dimension of the feature h and the $\sigma_{ci}$ and $\mu_{ci}$ distributions denote the standard deviation and expectation corresponding to the feature a under the category y = c.

Like other simple Bayesian classifiers, GaussianNB performs well on high-dimensional data. It is also friendly to missing values and can handle continuous numerical data well. GaussianNB requires less data for training, is a simple and efficient classifier with stable performance, and can be applied to a variety of different classification problems.

## 2.3. Genetic Algorithm

Genetic algorithms (GA) originated from computer simulations of biological systems and are a stochastic global search and optimization method that mimics the evolutionary process of organisms, simulating the natural selection and genetic reproduction, crossover, and mutation that occur in nature, which reproduce and evolve from generation to generation, and finally converge to a group of individuals best adapted to the environment to find a quality solution to the problem, which can be described by the following mathematical programming model [20]:

$$\begin{cases} maxf(X) & (1) \\ x \in R & (2) \\ R \subset U & (3) \end{cases}$$

In the above equation, where equation (1) is used as the objective function, X is the decision variable, equations (2) and (3) are used as constraints and R is a subset of the underlying space U. A feasible solution X is a solution consisting of solutions satisfying the constraints, and the set R of feasible solutions is the set consisting of solutions satisfying the constraints.

Common terms used in genetic algorithms are defined as follows:

Chromosomes: Chromosomes can also be referred to as genotyped individuals, a certain number of individuals make up a population, and the number of individuals in a population is called the population size.

Genes: Genes are elements of chromosomes that are used to represent an individual's characteristics. For example, if there is a chromosome S=1101, the four elements 1, 1, 0, 1 are called genes.

Adaptation: The degree to which each individual adapts to its environment is called the degree of adaptation. To capture the adaptive capacity of chromosomes, a function is introduced that measures each chromosome in the problem, called the fitness function. This function is often used to calculate the probability that an individual will be used in a population.

Genotype: Also known as genotype, the genome defines the genetic features and expressions that correspond to the chromosomes in the GA.

Phenotype: The characteristics of an organism's genotype as expressed in a given environment. The parameters corresponding to the decoded chromosomes in GA.

Each possible solution is encoded as a chromosome, and a large number of individuals make up the population. At each iteration of the population, the best individuals of the generation are selected based on their fitness and by performing replication, crossover, and mutation operations. During the evolution of each generation, the population becomes more and more adaptive, and eventually the optimal solution of the problem can be obtained by decoding the optimal individuals in the last generation, as show in algorithm 2. (Define: M: Population size; T: Iterations; $p_c$: Crossover probability; $p_m$: Mutation probability;)

```
Algorithm 2 Algorithm of Genetic Algorithm
1: Begin
2: initialize P(0);
3: t = 0
4: while t ≤ T do
5:     for i = 1 to M do
6:         Evaluate fitness of P(t);
7:     end for
8:     for i = 1 to M do
9:         Select operation to P(t);
10:    end for
11:    for i = 1 to M/2 do
12:        Crossover operation to P(t);
13:    end for
14:    for i = 1 to M do
15:        Mutation operation to P(t);
16:    end for
17:    for i = 1 to M do
18:        P(t + 1) = P(t)
19:    end for
20:    t = t + 1
21: end while
22: end
```

## 3. DATA AND FEATURE SELECTION

### 3.1. Data selection

The Backblaze website provides the public hard drive dataset, which was used as the experimental dataset in this paper. Because different hard disk manufacturers define and value their hard disk SMART characteristics differently, it is normal for the same SMART characteristics to have different meanings for different manufacturers' hard disks [21]. The datasets of numerous hard drive models exhibit a considerable proportion of normal hard drive information compared to a minimal number of failed hard drives, creating a very serious class imbalance and making hard drive failure prediction difficult. In this paper, the construction of the experimental dataset focuses on addressing the adverse effects caused by class imbalance. To ensure the validity of the study, the SMART data from Backblaze for the disk model ST4000DM000, which had the highest occurrence of failure states throughout the entire year of 2020, was utilized.

### 3.2. Feature selection

With the aim of enhancing classification accuracy, reduce training time and reduce overfitting, it is necessary to select the features that are more useful for predicting disc failure. There are 126 attributes on the SMART data disc. To avoid the challenges posed by large data volumes and numerous attributes, it becomes imperative to conduct feature selection. This process ensures that the dimensionality catastrophe is averted. Feature selection can remove unnecessary or redundant attributes, thus reducing the dimensionality of the feature space and the complexity of the model, which can help the model to better describe the data, enhance the modle's recognition rate and prediction ability, and reduce the training time and computational cost.

The attribute composition of the experimental dataset used in this study was determined based on a combination of the literature [22] and [23] where the discussion and experimental analysis were presented. Table 1 presents the selection of the 11 SMART attributes for the dataset in this paper.

Table 1. List of SMART data items selected for fault prediction.

| ID | SMART data item name |
| --- | --- |
| 1 | Raw Read Error Rate |
| 3 | Spin Up Time |
| 5 | Reallocated Sector Count |
| 5(raw) | Reallocated Sector Count(original value) |
| 7 | Seek Error Rate |
| 9 | Power On Hours |
| 187 | Reported Uncorrectable Error |
| 189 | High Fly Writes |
| 194 | Temperature Celsius |
| 197 | Current Pending Sector Count |
| 197(raw) | Current Pending Sector Count(original value) |

### 3.3. Data normalisation

Big data has characteristics such as large volume and variety, and large and disordered data can have a great impact on research results [24]. In machine learning tasks, large data should be pre-processed in a normalised way to improve the prediction accuracy of models [25,26] and achieve high performance of machine learning [27]. Data normalisation is an important step in data pre-processing. By normalising the data, it is possible to improve the robustness and performance of the algorithm by avoiding the dominant effect of highly variable attributes on the training of the model. Avoiding errors and ensuring that the data are in the same size space allows for a fairer comparison between attributes.

In this paper, StandarScaler is used to normalise the data so that the processed data conform to the standard normal distribution; then Min-Max normalisation is used to normalise the dataset (Equation follows) to map all the data to between [-1,1], which does not change the data distribution after processing the original data and better preserves the data characteristics.

$$X_{normal} = 2 * \frac{x - x_{min}}{x_{max} - x_{min}} - 1$$

The equation defines x as the pre-normalization value. $X_{normal}$ represents the normalized value, while $x_{min}$ and $x_{max}$ correspond to the minimum and maximum values of an attribute column in the dataset, respectively.

### 4. HYBRID POLICY SELECTION IMPLEMENTATION OF MULTIVARIATE GAN

To address the adverse effects of the imbalance of disk SMART data categories and the use of resampling to mitigate this problem, this paper makes improvements based on the data-level approach and implements a data-level hybrid policy selection method for disk failure prediction models based on multivariate GAN:

1. Based on three different GAN models for synthesising tabular data: CTGAN, CopulaGAN and CTAB-GAN, the three GAN models were trained separately using the real SMART dataset (the partitioned training set) to generate "fake data", and the synthesised tabular GAN models were filtered out for use in The disk failure samples synthesized by each table GAN model for balancing the disk SMART data were selected as the data set to be used to balance the real SMART data set. This is illustrated in Figure.3.

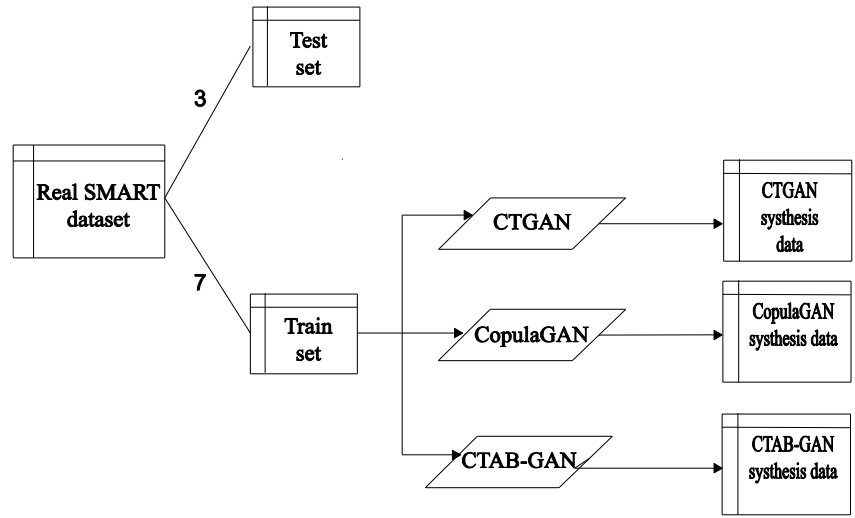

Figure 3.  Schematic of GAN synthetic dataset

2. The genetic algorithm populations were used to represent the proportion of mixed integration data from the three GAN models, and the proportion of mixed data was represented by the decoded phenotypes of the chromosomal genotypes. For example, Figure.4, the data sets from each GAN model were mixed and integrated into the real SMART dataset to build a positive and negative sample balanced dataset, and the classification model was used as a selection method to find the best data proportion for a particular classification model through generational gene duplication, exchange and mutation, using the "elite retention" method. The optimal ratio of synthetic GAN data mixed and integrated for a particular classification model is found, resulting in a positive and negative sample-balanced disc SMART dataset.

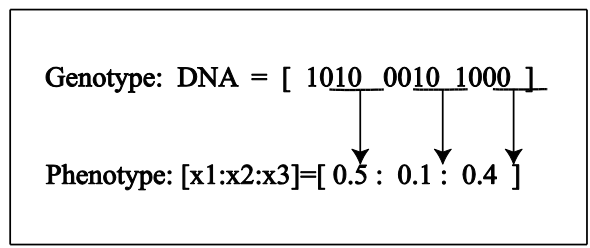

Figure 4. Illustration of genotype decoding into expressions

3. The implementation of the data-level hybrid policy selection method for multivariate GAN-based disk failure prediction models combines a classification model with a traditional genetic algorithm, incorporating the classification model as the fitness calculation function in the genetic algorithm, using the classification model as the fitness function in the genetic algorithm, so that the output of the trained predicted values of the classification model is used as the fitness of the chromosomal individuals (the hybrid integrated data portion of the synthetic data of the three GAN models).

After iterating the genetic algorithm, the best balanced disk SMART dataset for different classifiers is finally obtained, and then the balanced disk SAMRT dataset for this classifier is used on the specified classifier to achieve better results in disk failure prediction and improve the accuracy of disk failure prediction, and in this paper, the proposed data-level hybrid policy selection method is evaluated for validating the effectiveness of the multivariate GAN-based model in predicting disk failure.

## 5. EXPERIMENTATION AND ANALYSIS

Note: The parameter settings of the generative models (CTGAN, CopulaGAN, CTAB-GAN), classification models and genetic algorithms used in this experiment are listed in Table 2.

Table 2. Parameters involved in the experimental model.

| CTGAN | epochs=300 |
|---|---|
| CopulaGAN | epochs=300 |
| CTAB-GAN | Epochs=300 |
| MLP | input_features=11, hidden1=30, hidden2=30, out_features=2 |
| SVM | C=100, gamma=1 |
| Decision Tree | max_depth=9 |
| RandomForest | n_estimators=100 |
| GA | Population size=150, Iterations=50, Chromosome number=12, Crossover=0.8,Mutation=0.01 |

### 5.1. Experimental dataset setup

The base dataset used for this experiment was a 1:100 positive to negative sample ratio disc SMART dataset selected from a full year of a complete year (2020) of publicly available data published by Backblaze. In the test set, 30% of this dataset was chosen before and after balancing, while 70% was utilized as training set A to assess the classification effects before balancing, as the training set B for training each tabular GAN model to synthesize the data, and as the initial dataset C for mixing and integrating with each GAN model. The new balanced dataset is formed by mixing and integrating the data synthesized by each GAN model. A balanced dataset is constructed by integrating a mixture of disk failure data synthesized using a multivariate GAN into the base dataset as training data, and comparing the performance of the unbalanced training set A on the classification model.(Note: Training set A, training set B, and initial dataset C are equivalent and are 70% of the data split from the base dataset, and are set as A, B, and C to distinguish them from each other when later applied to different aspects.)

### 5.2. Evaluation indicators

The G-mean [6] was used to evaluate the performance of the classification model in this experiment, and a confusion matrix defined as follows (see Table 3) was used to calculate the G-mean values.

Table 3. Confusion matrix.

| **Predicted as failure** | **Predicted as normal** | **Classified as** |
|---|---|---|
| TP(True Positive) | FN(False Negative) | True Fault |
| FP(False Positive) | TN(True Negative) | True Normal |

The overall performance of the classifier is evaluated using the G-mean (geometric mean). An elevated G-mean is achieved whenever both bad and normal discs are classified correctly, and the use of other evaluation metrics is not accurate for classifying unbalanced data on the classifier:

### 5.3. Experimental setup

#### 5.3.1. GA Advantage Search Experiment

The experimental structure of the data-level hybrid policy selection method based on multivariate GAN for the disk failure prediction model is shown in Figure.5.

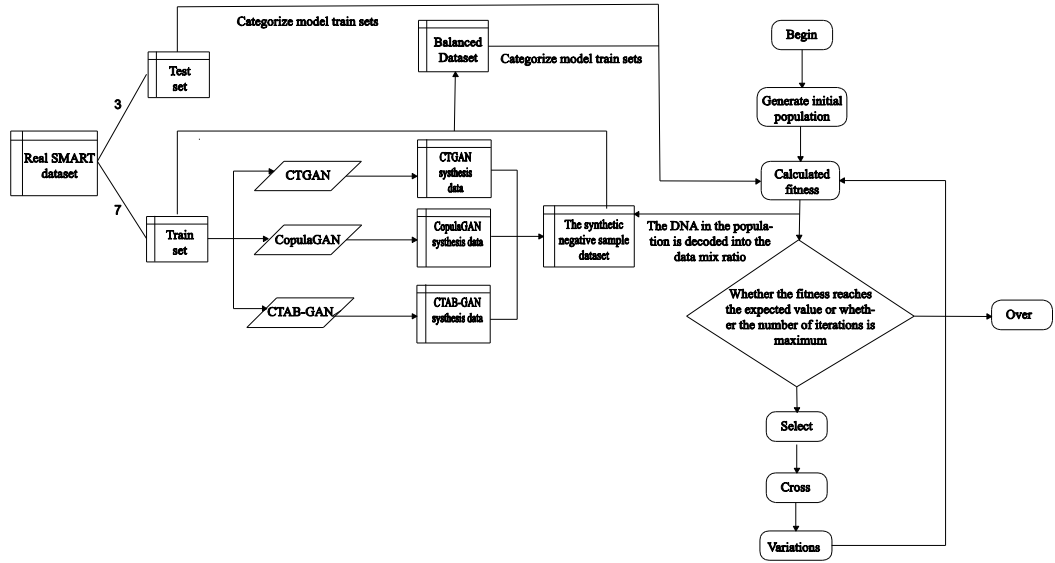

Figure 5. Experimental structure

#### 5.3.2. Validation experiment

As shown above, the training set B, divided by the base dataset, is used to train generative models: CTGAN, CopulaGAN and CTAB-GAN, and the disk failure samples synthesized by each tabular GAN model for balancing the disk SMART data are filtered out and used as the data set for balancing the real SMART data.

The genetic algorithm is set with an appropriate population size N, and the number of chromosome genes n. During the iterative process of the genetic algorithm, the three GAN-synthesized datasets are mixed and integrated into the initial dataset C. The mixing ratio is determined by the decoded chromosome genotypes. The resulting balanced dataset is then utilized as the training set to train the classification model, which serves as an adaptation function. The predicted output of the classification model is used as the fitness measure for each chromosome. After iterating until the termination condition of the genetic algorithm is reached, a dataset is generated that achieves the best training and testing results in terms of the integration ratio of multivariate GAN synthetic data mixtures, while simultaneously achieving a balance between both positive and negative data instances.

In this validation experiment, three sets of experiments are set up to verify the effectiveness of the data-level hybrid policy selection approach proposed in this paper for the multivariate GAN based disk failure prediction model, using different training sets on multiple classification models. The different training sets used in the three sets of experiments are: a: The base dataset is divided, with 70% allocated to the training set; b: Using CTGAN, CopulaGAN and CTAB-GAN to synthesize a balanced dataset together with the initial dataset; c: The dataset c is a mixture of synthetic data from the multivariate GAN and added to the initial dataset to form a balanced dataset based on the results of the genetic algorithm on a specific classification model. the data from the multivariate GAN is mixed and integrated into the initial data set to form the balanced dataset. As illustrated in Figure.3, Utillizing the training set A as the benchmark dataset, Table 4 illustrates the distribution and ratio of both positive and negative data instances within the training set for all three sets of comparative experiments.

Table 4. Proportion of both positive and negative data instances from the experimental dataset.

| Dataset Composition | Proportion of positive and negative samples |
|---|---|
| Data ratio of real data | 172：17400 |
| Balanced data(CTGAN or CopulaGAN or CTAB-GAN) | 17400：17400 |
| Balanced data(CTGAN+ CopulaGAN+CTAB-GAN) | 17400:17400 |

## 5.4. Comparison of experimental results

### 5.4.1. Comparison of experimental results

The results of the GA search experiments can be found in Table 5. By examining the data presented in the aforementioned Table, it becomes evident that for the best prediction of disk faults using different classification models, the GA-seeking ratio of the hybrid ensemble of multivariate GAN synthetic data is different. In this paper, we investigate the data-level hybrid policy selection for a multivariate GAN based model of disk failure prediction, which positively contributes to the prediction of disk failures.

Table 5. Optimal data mixing ratios for GA search.

| Classification Model | The ratio optimized by GA |
|---|---|
| MLP | 0.412: 0.0:0.588 |
| SVM | 0.857:0.0:0.143 |
| Decision Tree | 0.40:0.133:0.467 |
| Bayes | 0.565:0.348:0.087 |
| RandomForest | 0.167:0.167:0.669 |

## 5.5. Validating experimental results

The three datasets a, b, c used for comparison and validation experiments are used as training sets to train different classification models and output G-mean results for disk fault prediction. The third dataset, c, used for the unclassified model, corresponds to the balance of positive and negative samples constructed by combining the results of GA with different classification models and the initial dataset c. The experimental results are shown in Table 6.

From the results of the validation experiment it can be seen that compared with the other four datasets, the classification model trained on training set A (With a ratio of 1:100 between both positive and negative data instances ) has the lowest disk failure prediction results. The reason is that the training set A, a real disk SMART dataset, has very unbalanced positive to negative samples, which causes the classification model to focus on the majority class, the healthy samples, and ignore the influence of the minority class, the faulty samples. A balanced dataset of both positive and negative data instances constructed from the synthetic dataset produced by the single-table GAN mode and the initial dataset C, is used to train the classification model and achieve better results. However, the balanced dataset constructed by using the optimal mixture of GA and classification models to find the optimal integration ratio of the multivariate GAN synthesized data outperforms the balanced dataset with a single tabular GAN on all five classification models, its G-mean output on the above five classification models is 86.4% on average.

Table 6. Test results of classification models trained with different training sets.

| Classification Model | 1:100 | CTGAN | CopulaGAN | CTAB-GAN | Optimized by GA |
|---|---|---|---|---|---|
| ANN | 0.426 | 0.858 | 0.657 | 0.711 | 0.902 |
| SVM | 0.426 | 0.852 | 0.621 | 0.698 | 0.854 |
| Decision Tree | 0.603 | 0.543 | 0.670 | 0.855 | 0.904 |
| Bayes | 0.654 | 0.045 | 0.090 | 0.619 | 0.725 |
| RandomForest | 0.452 | 0.657 | 0.564 | 0.896 | 0.936 |

It can be seen that the effectiveness of the data-level hybrid policy selection employed in the multivariate GAN-based disk failure prediction model investigated in this study has been verified and beneficial for disk failure prediction efforts.

## 6. SUMMARY AND PROSPECT

In the era of extensive data growth, particularly within the realm of big data, the storage of this data primarily takes place on disks located in data centers. However, in the unfortunate event of disk damage, the data stored on these disks faces the risk of permanent loss. Hence, the prediction of disk failure becomes crucial in ensuring data integrity. Nonetheless, an inherent challenge lies in the severe imbalance between fault samples and normal samples within the SMART data that records the characteristics of these disks. This severe imbalance poses significant inconvenience in conducting accurate disk failure prediction. The main contributions of this paper are the use of hybrid ensemble multivariate GAN synthesized data combined with genetic algorithms to construct balanced datasets at the data level, and the experimental validation of the performance of pre- and post-balanced datasets on various classification models.

While the suggested approach of employing a balanced dataset method does enhance the accuracy of disk failure prediction to a certain degree, there remains scope for further improvement. Key areas for consideration encompass the following aspects:

1. Feature selection involves determining the significance of attributes within a dataset through a selection process. If a smaller dimensional dataset can be constructed that better reflects the disk properties, it will improve the efficiency of disk failure prediction.

2. The synthetic datasets used to balance the datasets in this paper are directly drawn from the sample derived from the GAN's generation. The effectiveness can be enhanced by eliminating noise from the data and ensuring a higher similarity between the synthetic data sample and the real data sample through the process of data cleaning.

3. When training data with classification models, the goal is to learn a stable model that performs well in all aspects, but in practice this is not the case and sometimes only multiple models with preferences can be obtained. Integrated learning can be used to bring together multiple classification methods to improve classification accuracy, so combining ensemble learning with evolutionary learning for specific disk failure prediction tasks is a good direction for research.

## ACKNOWLEDGEMENT

The research work in this paper was supported by the Shandong Provincial Natural Science Foundation of China (Grant No. ZR2019LZH003), Science and Technology Plan Project of University of Jinan (No. XKY2078) and Teaching Research Project of University of Jinan (No. J2158). Yuehui Chen is the author to whom all correspondence should be addressed.

**Authors**

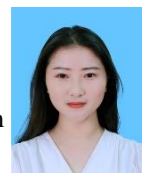

**Shuangshuang Yuan** Currently, she is studying for a master's degree in Computer science from University of Jinan. Her main research interests are intelligent storage, data processing, and the application of deep learning.

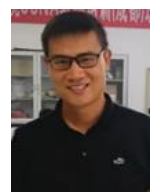

**Peng Wu** received Master from University of Jinan and a doctoral degree from Beijing Normal University. Currently, His main research interests are pattern recognition, bioinformatics, intelligent computing theory and application research.

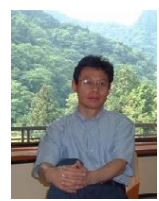

**Yuehui Chen**  received his Ph.D. in Computer and Electronic Engineering from Kumamoto

University, Japan. currently, His main research interests are theory and application of intelligent computing, bioinformatics and systems biology.